\documentclass[review]{elsarticle}

\usepackage{amsmath,amssymb,amsfonts,yfonts,graphicx,epstopdf,epsfig,url}
\usepackage[margin=1in]{geometry}
\usepackage{url,hyperref}
\usepackage[justification=centering]{caption}
\usepackage{formatfig}
\usepackage{lettrine}
\usepackage{color}
\usepackage{lineno,hyperref}
\modulolinenumbers[5]

\journal{Elsevier Journal of Signal Processing}

\newcommand{\order}[1]{$\mathcal{O}(#1)$}

\begin{document}

\begin{frontmatter}

\title{\sc Optimizing Codes for Source Separation in Color Image Demosaicing and Compressive Video Recovery}

\author[mymainaddress]{Alankar Kotwal}
\ead{alankar.kotwal@iitb.ac.in}
\address[mymainaddress]{Department of Electrical Engineering, IIT Bombay}

\author[mymainaddress2]{Ajit Rajwade\corref{mycorrespondingauthor}}
\cortext[mycorrespondingauthor]{Corresponding author}
\ead{ajitvr@cse.iitb.ac.in}
\address[mymainaddress2]{Department of Computer Science and Engineering, IIT Bombay}
\fntext[mymainaddress]{AK gratefully acknowledges support from the Electrical Engineering department at IIT Bombay.}
\fntext[mymainaddress2]{AR gratefully acknowledges support from IIT Bombay Seed Grant number 14IRCCSG012.}

\begin{abstract}
There exist several applications in image processing (eg: video compressed sensing \cite{Hitomi2011} and color image demosaicing \cite{MoghadamAKR13}) which require separation of constituent images given measurements in the form of a coded superposition of those images. Physically practical code patterns in these applications are non-negative, systematically structured, and do not always obey the nice incoherence properties of other patterns such as Gaussian codes, which can adversely affect reconstruction performance. The contribution of this paper is to design code patterns for video compressed sensing and demosaicing by minimizing the mutual coherence of the matrix $\boldsymbol{\Phi \Psi}$ where $\boldsymbol{\Phi}$ represents the sensing matrix created from the code, and $\boldsymbol{\Psi}$ is the signal representation matrix. Our main contribution is that we explicitly take into account the special structure of those code patterns as required by these applications: (1)~non-negativity, (2)~block-diagonal nature, and (3)~circular shifting. In particular, the last property enables for accurate and seamless patch-wise reconstruction for some important compressed sensing architectures.
\end{abstract}


\end{frontmatter}

\section{\sc Introduction and Related Work}
\lettrine[lines=2]{\gothfamily \scalebox{1.5}{C}}{ompressed} sensing has been explored as an alternative way of sampling continuous-time signals. Its success with still images has inspired efforts to apply it to video. Indeed,~\cite{Hitomi2011,ReddyVC11} achieve compression across time by combining frames into coded snapshots while sensing and separating them with a pre-trained over-complete dictionary. \cite{Hitomi2011,ReddyVC11} make a diagonal choice for the sensing matrix for each frame. $T$ input frames of size $N_1 \times N_2$ denoted as $\{\boldsymbol{X_i}\}_{i=1}^{T}$ are sensed so that output $\boldsymbol{Y}$ of size $N_1 \times N_2$ appears as a pixel-wise coded superposition (dictated by the sensing matrices $\boldsymbol{\Phi_i}$) of the input frames. The pixel-wise codes represent modulation of incident light by patterns as per a pre-programmed DMD (digital micromirror device) or LCoS (liquid crystal on silicon) device. $\boldsymbol{Y}$ is sometimes termed the coded snapshot image. The sensing framework is
\begin{equation}
\textrm{vec}(\boldsymbol{Y}) = \sum_{i=1}^{T} \boldsymbol{\Phi_i} \textrm{vec}(\boldsymbol{X_i}) = \boldsymbol{\Phi} \textrm{vec} (\boldsymbol{X}).
\label{eq:hitSensing}
\end{equation}
Here, each $\boldsymbol{\Phi_i}$ is a $n \times n$ (where $n = N_1 N_2$ is the number of pixels per frame) non-negative (possibly, but not necessarily binary) diagonal sensing matrix with vectorized code elements on the diagonal, and the overall sensing matrix $\boldsymbol{\Phi} = (\boldsymbol{\Phi_1} | \boldsymbol{\Phi_2} | ... \boldsymbol{\Phi_T})$ has size $n \times nT$. The complete $N_1 \times N_2 \times T$ video is represented as $\boldsymbol{X} = (\boldsymbol{X_1} | \boldsymbol{X_2} | ... | \boldsymbol{X_T})$. This acquisition model also holds in color image demosaicing with $T = 3$.

The sparsifying basis here could be a 3D dictionary learned on video patches or a universal basis such as 3D-DCT (discrete cosine transform). Given this dictionary, called $\boldsymbol{D}$, any given signal $\boldsymbol{X}$, and in particular, its frames $\{\boldsymbol{X_i}\}_{i=1}^{T}$ can be approximately reconstructed as a sum of its projections $\boldsymbol{\alpha_j}$ on the $K$ atoms in $\boldsymbol{D}$:
\begin{equation}
\textrm{vec}(\boldsymbol{X_i}) = \sum_{j=1}^{K} \boldsymbol{D_{ji} \alpha_{j}}
\label{eq:hitDict}
\end{equation}
where $\boldsymbol{D_{ji}}$ is the $i^\text{th}$ frame in the $j^\text{th}$ 3-D dictionary atom $\boldsymbol{D_{ji}}$. From the measurements and the dictionary, the input images are recovered solving the following optimization problem:
\begin{equation}
\min_{\boldsymbol{\alpha}} \|\boldsymbol{\alpha}\|_0 \text{ subject to } \left\| \textrm{vec}(\boldsymbol{Y}) - \sum_{i=1}^{T} \boldsymbol{\Phi_i} \sum_{j=1}^{K} \boldsymbol{D_{ji} \alpha_{j}} \right\|_2 \leq \epsilon.
\label{eq:hitOpt}
\end{equation}
This problem can be approximately solved with sparse recovery techniques like orthogonal matching pursuit \cite{Cai2011}. The drawback here, though, is that the 3D dictionary imposes a smoothness assumption on the scene. Since a linear combination of dictionary atoms cannot speed an atom up, the typical speeds of objects moving in the video must be roughly the same as the dictionary. Also, because of the nature of the training data, the dictionary fails to sparsely represent sudden scene changes caused by, say, lighting or occlusion. Other techniques like \cite{VR201104} exploit additional structure within the signal, like periodicity, rigid motion or analytical motion models and cannot be used in the general video sensing case.

We try relaxing these constraints using a source-separation approach~\cite{Studer201412}, where precise error bounds on the recovery of the images have been derived, with possible improvement using the techniques in~\cite{Cai2010}. Each of the coded snapshots is treated as a mixture of sources, each sparse in some basis. We experimented with basis pursuit recovery with Gaussian-random sensing matrices, getting excellent results with no visible ghosting for both similar and radically different images. Unfortunately, the actually realizable positive sensing matrices are more structured, and do not have the nice incoherence properties of Gaussian-random matrices, which are sufficient conditions for near-accurate recovery as derived in~\cite{Studer201412}.

\textbf{Survey of Previous Work:} Here, we aim to design such \emph{structured} sensing matrices with low mutual coherence (see Section \ref{subsec:coh} for a precise definition), making them ideal for compressed video. The existing literature on designing compressed sensing matrices can be broadly divided into two categories: (1) optimizing mutual coherence and (2) information theoretic approaches. Techniques in the former category include \cite{Li2017,Pereira2014,Abolghasemi2012,Duarte200907, Elad200610, Mordechay2014,Hong2016,XLi2017}. Most approaches \cite{Abolghasemi2012,Mordechay2014,Elad200610,Duarte200907,Hong2016,XLi2017} involve minimizing the difference between a Gram matrix and the identity matrix, thereby minimizing some forms of an average of normalized dot products of effective dictionary columns. Although minimizing averages doesn't guarantee minimizing the maximum (which is the mutual coherence in this case) of the quantities forming this average, the techniques have been quite effective. Methods such as \cite{Li2017} first estimate signal support using matrices designed to minimize coherence, followed by matrices designed to reduce MSE on the known support. \cite{Pereira2014} specializes to orthogonal or biorthogonal bases. However these methods are designed for general compressed sensing matrices and do not account for the \emph{special structure} of the sensing matrices used for video compressed sensing or for demosaicing, a framework which this paper expressly deals with. Techniques in the second category include \cite{Carson2012,Renna2013,Weiss2007}. These papers design sensing matrices $\boldsymbol{\Phi}$ such that the mutual information between a set of small patches $\{\boldsymbol{x_i}\}_{i=1}^{N_p}$ and their corresponding projections $\{\boldsymbol{y_i}\}_{i=1}^{N_p}$ where $\boldsymbol{y_i} = \boldsymbol{\Phi x_i}$, is maximized. Computing this mutual information first requires estimation of the probability density function of the patches and their projections, using Gaussian mixture models for instance. This can be expensive and is an iterative process. Moreover these learned GMMs for a class of patches may not be general enough. 

\textbf{Contributions:} Quite importantly, the aforementioned techniques in both categories do not account for the special structure of the sensing matrices described for video compressed sensing \cite{Hitomi2011,ReddyVC11} and demosaicing \cite{Hirakawa2008,MoghadamAKR13}, which is the primary aim of this paper. We design codes to enable the reconstruction of small patches from $\boldsymbol{Y}$ in the model described in Eq. \ref{eq:hitSensing} using algorithms such as basis pursuit. This is taking into account the fact that although the acquisition of $\boldsymbol{Y}$ is performed at the level of a group of $T$ consecutive frames (and specifically, \emph{not} at the level of individual patches), the reconstruction can in fact proceed patch-wise as shown in \cite{Hitomi2011}, and as we shall show in Section \ref{subsec:circ_sym}. Hence the sensing matrix can in fact be designed at the level of small patches, followed by a \emph{tiling} to create the complete sensing matrix. We also demonstrate the benefit of designing patch-based codes whose \emph{circular shifts} also have low mutual coherence. Reconstruction results are presented for video compressive recovery and color image demosaicing.

\textbf{Organization:} The complete theoretical development is presented in Section \ref{sec:method_analysis}, followed by experimental results on video compressed sensing and color image demosaicing in Section \ref{sec:results}, followed by a discussion in Section \ref{sec:concl}.

\section{\sc Method and Analysis}
\label{sec:method_analysis}

\subsection{Our framework}
We propose to use a recovery method different from the one used in \cite{Hitomi2011}, within the same acquisition framework. Thus, our signals are still acquired according to Eq.~\ref{eq:hitSensing}. However, the choice of the sparsifying basis is different: we use a 2D-DCT basis $\boldsymbol{D}$ to model each frame in the input data, though our method is general enough to work for any other basis. The dictionary $\boldsymbol{\Psi}$ sparsifying the entire video sequence, thus, is a block-diagonal matrix with the $n \times n$ sparsifying basis $\boldsymbol{D}$ on the diagonal where $n = N_1 N_2$ is the number of pixels per video frame. Thus,
\begin{align}
\textrm{vec}(\boldsymbol{Y}) &= \begin{pmatrix}
\boldsymbol{\Phi_1} & \hdots & \boldsymbol{\Phi_T}
\end{pmatrix}
\begin{pmatrix}
\boldsymbol{D \alpha_1} &
\hdots &
\boldsymbol{D \alpha_T}
\end{pmatrix}^T \\
&= \begin{pmatrix}
\boldsymbol{\Phi_1 D} & \hdots & \boldsymbol{\Phi_T D}
\end{pmatrix}
\begin{pmatrix}
\boldsymbol{\alpha_1} &
\hdots &
\boldsymbol{\alpha_T}
\end{pmatrix}^T
\label{eq:sourceSepModel}
\end{align}

\noindent Given a measurement $\boldsymbol{Y}$, we recover the input $\{\boldsymbol{X_i}\}_{i=1}^{T}$ through the DCT coefficients $\boldsymbol{\alpha}$ by solving the optimization problem
\begin{equation}
\min_{\boldsymbol{\alpha}} \|\boldsymbol{\alpha}\|_1 \text{ subject to } \|\textrm{vec}(\boldsymbol{Y}) - \boldsymbol{\Phi \Psi \alpha}\|_2 \leq \epsilon,\ \boldsymbol{\alpha} = \begin{pmatrix}
\boldsymbol{\alpha_1} &
\boldsymbol{\alpha_2} &
\hdots &
\boldsymbol{\alpha_T}
\end{pmatrix}^T.
\label{eq:sourceSepOpt}
\end{equation}
In our implementation we used the \texttt{CVX}~\cite{cvx} solver for solving the convex optimization problem in Eq.~\ref{eq:sourceSepOpt}.

\subsection{Calculation of mutual coherence for our matrices}
\label{subsec:coh}
Our aim, then, is to optimize the sensing matrices $\boldsymbol{\Phi_i}$ directly for minimum mutual coherence with gradient descent. The criterion of coherence is motivated by the fact that the error bounds on signal recovery presented in \cite{Studer201412} are monotonically increasing functions of coherence. We now calculate gradients of the mutual coherence with respect to the (non-zero, \textit{i.e.}, diagonal) elements of $\boldsymbol{\Phi_i}$. As in Eq.~\ref{eq:sourceSepModel}, with an $n \times n$ dictionary $\boldsymbol{D}$ for representing a single 2D frame, we have the effective dictionary
\begin{align}
\boldsymbol{\Phi \Psi} = \begin{pmatrix}
\boldsymbol{\Phi_1 D} & \boldsymbol{\Phi_2 D} & \hdots & \boldsymbol{\Phi_T D}
\end{pmatrix}.
\end{align}
The mutual coherence of a general matrix $\boldsymbol{H}$, with the $i^\text{th}$ column defined as $\boldsymbol{h_i}$, is 
\begin{align}
\mu (\boldsymbol{H}) = \max_{i \neq j} \frac{|\left< \boldsymbol{h_i}, \boldsymbol{h_j} \right>|}{\sqrt{\left< \boldsymbol{h_i}, \boldsymbol{h_i} \right>\left< \boldsymbol{h_j}, \boldsymbol{h_j} \right>}}.
\label{eq:mu_coh}
\end{align}
This expression contains \texttt{max} and \texttt{abs} functions that a gradient-based scheme cannot handle. Instead, we soften the \texttt{max} and convert the \texttt{abs} to a square by using the following approximation, which is very accurate for large enough $\theta$,
\begin{align}
\max_i \{t_i^2\}_{i=1}^{n} \approx \frac{1}{\theta} \log \sum_{i=1}^{n} e^{\theta t_i^2}.
\label{eq:softMax}
\end{align} 

We need to evaluate the mutual coherence of the dictionary $\boldsymbol{\Phi \Psi}$ as a function of the non-zero elements of $\boldsymbol{\Phi}$. We will call the time index varying from $1$ to $T$ as $\mu$ or $\nu$, and the spatial index varying from $1$ to $n$ as $\alpha$, $\beta$ or $\gamma$. The $\mu^\text{th}$ block of $\boldsymbol{\Phi}$ is thus $\boldsymbol{\Phi_\mu}$. Let the $\beta^\text{th}$ diagonal element of $\boldsymbol{\Phi_\mu}$ be $\phi_{\mu\beta}$. Define the $\alpha^\text{th}$ column of $\boldsymbol{D}^T$ to be $\boldsymbol{d_\alpha}$. Then, it can be shown~[\ref{App:derCoh}] that the normalized dot product between the $\beta^\text{th}$ column of the $\mu^\text{th}$ block and the $\gamma^\text{th}$ column of the $\nu^\text{th}$ block is
\begin{align}
M_{\mu \nu}(\beta\gamma) &= \frac{\sum_{\alpha = 1}^{n} \phi_{\mu \alpha} \phi_{\nu \alpha} d_\alpha (\beta) d_\alpha (\gamma)}{\sqrt{\left( \sum_{\alpha = 1}^{n} \phi_{\mu \alpha}^2 d^2_\alpha (\beta) \right) \left( \sum_{\tau = 1}^{n} \phi_{\nu \tau}^2 d^2_\tau (\gamma) \right)}}.
\end{align}

Finally, using the squared soft-max function [Eq.~\ref{eq:softMax}] to deal with the \texttt{max} and the \texttt{abs} in the mutual coherence expression, we get the squared soft mutual coherence $\mathcal{C}$ to be
\begin{align}
\mathcal{C} = \frac{1}{\theta} \log \left[ \sum_{\mu=1}^{T} \sum_{\nu=1}^{\mu-1} \sum_{\beta=1}^{n} \sum_{\gamma=1}^{n} e^{\theta M_{\mu \nu}^2(\beta\gamma)} + \sum_{\mu=1}^{T} \sum_{\beta=1}^{n} \sum_{\gamma=1}^{\beta - 1} e^{\theta M_{\mu \mu}^2(\beta\gamma)} \right].
\end{align}

In the above, the first term corresponds to all ($\mu > \nu$) blocks that are below the block diagonal. Here, we consider all terms in the given block for the maximum. The second term corresponds to ($\mu = \nu$) blocks on the block diagonal. Here, we consider only consider ($\beta > \gamma$) below-diagonal elements for the maximum.

Note that lowering the value of $\theta$ in Eq \ref{eq:softMax} will take us closer toward an average of the normalized dot-product values (often referred to as `average coherence') instead of the mutual coherence defined in Eq. \ref{eq:softMax} and Eq. \ref{eq:mu_coh}. While this approach has been adopted in many papers, we observed equally good results with a large $\theta$ and with a much faster optimization. As such, our method remains unchanged even with smaller values of $\theta$.

\subsection{Calculation of coherence derivatives}
We note that the $\mathcal{C}$ computed in the section above is a function of $\boldsymbol{\Phi}$. We differentiate $\mathcal{C}$ with respect to $\Phi_{\delta \epsilon}$, the elements located on the diagonals of each of the square components, i.e. on the diagonals of $\boldsymbol{\Phi_1},\boldsymbol{\Phi_2},...,\boldsymbol{\Phi_T}$. These are in some sense, the $nT$ `degrees of freedom' of the $n \times nT$ matrix $\boldsymbol{\Phi}$. For this, we define the numerator of the expression for $M_{\mu \nu}(\beta\gamma)$ as $\chi_{\mu \nu}(\beta\gamma)$ and the denominator as $\xi_{\mu \nu}(\beta\gamma)$. The derivative of the objective function can be found in terms of these quantities. Defining $\uparrow_{\mu \delta}$ to be the Kronecker delta function that is 1 if and only if $\mu = \delta$, it can be shown that~[\ref{App:derCohDer}]

\begin{equation}
\frac{d\chi_{\mu \nu}(\beta\gamma)}{d\phi_{\delta \epsilon}} = d_\epsilon (\beta) d_\epsilon (\gamma) \left( \phi_{\mu \epsilon} \uparrow_{\nu \delta} + \uparrow_{\mu \delta} \phi_{\nu \epsilon} \right)
\end{equation}
\begin{equation}
\frac{d\xi_{\mu \nu}(\beta\gamma)}{d\phi_{\delta \epsilon}} = \frac{1}{\xi_{\mu \nu}(\beta \gamma)} \left[ \phi_{\mu \epsilon} d_\epsilon^2(\beta) \uparrow_{\mu \delta} \sum_{\tau = 1}^{n} \phi_{\nu \tau}^2 d^2_\tau (\gamma) + \phi_{\nu \epsilon} d_\epsilon^2(\gamma) \uparrow_{\nu \delta} \sum_{\alpha = 1}^{n} \phi_{\mu \alpha}^2 d^2_\alpha (\beta) \right].
\end{equation}

\noindent Using these, we perform projected gradient descent (to maintain non-negativity of $\boldsymbol{\Phi}$) with adaptive step-size and use a multi-start strategy to combat the non-convexity of the problem. We settle for the matrix produced by the start that yielded the least value of $\mathcal{C}$. In practice, we needed a smaller number ($\sim 20$) of starts.

For reasons to become clear further in this paper, we term matrices designed this way as `non-circularly designed matrices'.

\subsection{Time complexity and the need for something more}
The calculation of coherence for a matrix requires us to evaluate normalized dot products between columns of the matrix. In our case, the size of the matrix is $n \times nT$, and each dot product needs \order{n} operations, warranting the calculation of \order{n^3 T^2} quantities. Optimizing this rapidly becomes intractable as $n$ increases. The performance of gradient descent on this non-convex optimization problem also worsens as the dimensionality of the search-space (\order{nT}) increases. Empirically, we observe that it is intractable to design codes that are more than $20 \times 20$ in size in any reasonable time, whereas the size of typical images in video compressed sensing or demosaicing problems is far higher. This points to the fact that we need something more to make designing effective codes possible.

\subsection{Circularly-symmetric coherence minimization}
\label{subsec:circ_sym}
An important aspect of the compressive architecture described in Eq.~\ref{eq:hitSensing} is as follows. On one hand, the compressive measurement corresponds to a sequence of \emph{entire} video frames producing a coded snapshot image $\boldsymbol{Y}$ of size $N_1 \times N_2$. On the other hand however, the reconstruction can actually be performed in a patch-wise manner across overlapping patches in a sliding window fashion, as described in \cite{Hitomi2011} (or as in \cite{Rajwade2013}, for a related compressive architecture in hyperspectral imaging). This is because for a $m \times m$ patch $\boldsymbol{y_i}$ ($m^2 \ll n$) at location $i$ in $\boldsymbol{Y}$, we have
\begin{equation}
\textrm{vec}(\boldsymbol{y_i}) = \sum_{t=1}^T \boldsymbol{\phi_{it}} \textrm{vec}(\boldsymbol{x_{it}}) = \sum_{t=1}^T \boldsymbol{\phi_{it} \tilde{D}_{m^2 \times K} \alpha_{it}}, 
\end{equation}
where $\boldsymbol{\phi_{it}}$ is a $m^2 \times m^2$ diagonal sub-matrix of $\boldsymbol{\Phi_t}$ representing the portion of the code that modulates the pixels inside an $m \times m$ patch around location $i$ in frame $t$, and $\boldsymbol{x_{it}}$ represents a $m \times m$ patch around location $i$ in frame $\boldsymbol{X_t}$. Exploiting the fact that each patch $\boldsymbol{x_{it}}$ will be sparse in some dictionary $\boldsymbol{\tilde{D}}_{m^2 \times K}$, its dictionary codes $\boldsymbol{\alpha_{it}}$ (and hence the patch itself) can be independently inferred for each $i$, by solving the following optimization problem:
\begin{equation}
\textrm{min} \|\boldsymbol{\alpha_i}\|_1 \textrm{ subject to } \|\textrm{vec}(\boldsymbol{y_i})-\boldsymbol{\phi_i \tilde{D} \alpha_i}\|_2 \leq \epsilon
\label{eq:BP_patchwise}
\end{equation}
where $\boldsymbol{\phi_i} = (\boldsymbol{\phi_{i1}} | \boldsymbol{\phi_{i2}} | ... | \boldsymbol{\phi_{iT}})$ has size $m^2 \times m^2T$,$\boldsymbol{\alpha_i} = (\boldsymbol{\alpha_{i1}} | \boldsymbol{\alpha_{i2}} | ...| \boldsymbol{\alpha_{iT}})$ has size $TK \times 1$, and $\boldsymbol{\tilde{D}} \in \mathbb{R}^{m^2 \times KT}$ is a dictionary with $KT$ columns obtained by concatenating identical dictionaries $\boldsymbol{\tilde{D}}_{m^2 \times K}$ with $K$ columns.
The reconstruction of $\boldsymbol{x_{i}} = \boldsymbol{\tilde{D} \alpha_i}$ is then followed by an averaging operation in pixels that are covered by more than one patch. This avoids patch-seam artifacts that would occur if only non-overlapping patches were reconstructed.

\begin{figure}
\centering
\includegraphics[scale=0.5]{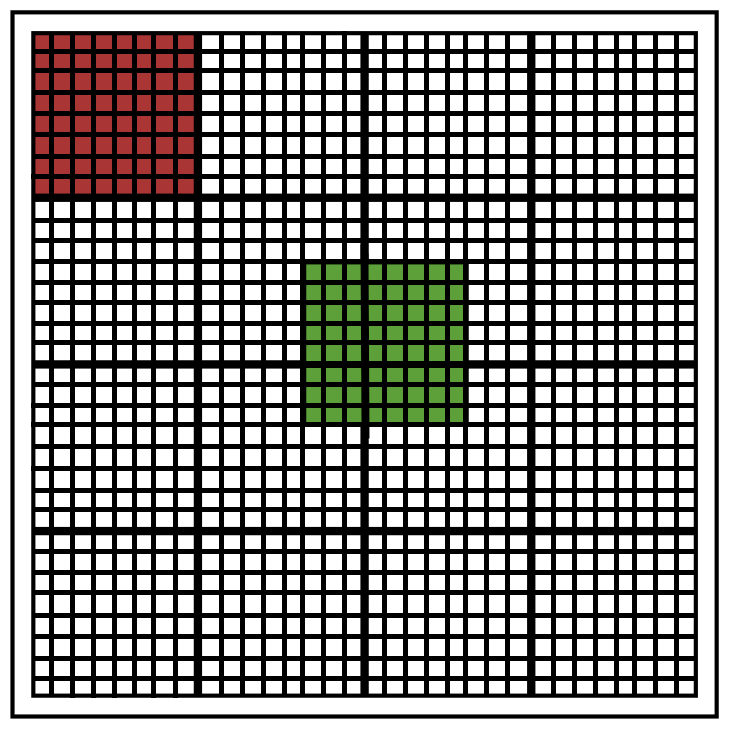}
\caption{Motivation behind circularly-shifted optimization. The code-mask for the green patch is not aligned with the usual patch boundaries unlike the red patch, and is in fact a circular shift of the original patch-aligned code-mask (see text for more details).}
\label{fig:circMot}
\end{figure}

As a matter of fact, this leads us to designing smaller code-masks (of size $m \times m$ per frame, where $m^2 \ll n$) and tiling them to fit the image size we are dealing with. This helps us get around the issue of computational tractability. A small coherence for the designed patch guarantees good reconstruction for patches of size $m \times m$ exactly aligned with the code-mask; however, other patches see a code that is a circular shift of the original code. Fig~\ref{fig:circMot} provides a visual explanation. The big outer square denotes the image. On top of the image we show tiled designed codes. Now, the patch in red clearly multiplies with the exact designed code; however the patch in green multiplies with a code that is a version of the original code but shifted in both the coordinates circularly. Of course, one could seek to reconstruct only those patches that are aligned exactly with the code mask, but that will produce patch-seam artifacts and require heuristic post-processing operations like deblocking.

This points to designing sensing matrices that have small coherence in all their circular permutations. Note that these permutations happen in two dimensions and must be handled as such. To this end, we modify the above objective function to minimize the maximum coherence resulting from all circularly-shifted vectorized versions of $\boldsymbol{\Phi}$. We thus have
\begin{align}
\mathcal{C} = \frac{1}{\theta} \log \left[ \sum_{\zeta \in \text{perm}(\Phi)} \left[ \sum_{\mu=1}^{T} \sum_{\nu=1}^{\mu-1} \sum_{\beta=1}^{n} \sum_{\gamma=1}^{n} e^{\theta M_{\mu \nu}^{(\zeta)2}(\beta\gamma)} + \sum_{\mu=1}^{T} \sum_{\beta=1}^{n} \sum_{\gamma=1}^{\beta - 1} e^{\theta M_{\mu \mu}^{(\zeta)2} (\beta\gamma)} \right] \right]
\end{align}
where $M_{\mu \nu}^{(\zeta)}(\beta\gamma)$ represents the normalized dot product between the $\beta^\text{th}$ column of the $\mu^\text{th}$ block and the $\gamma^\text{th}$ column of the $\nu^\text{th}$ block, resulting from the instance of the circular permutation $\zeta$ of $\Phi$. Derivatives of this expression are found exactly like in \ref{App:derCohDer}, except that the $\mu$, $\nu$, $\beta$ and $\gamma$ parameters are now subjected to the appropriate circular permutation. Matrices designed this way are henceforth referred to as `circularly designed matrices'

For $m \times m$ patches, the time complexity for determining this maximum coherence among all circular permutations is \order{\tilde{m}^4 T^2} where $\tilde{m} = m \times m$, out of which a \order{\tilde{m}^3 T^2} term arises from the calculation of coherence for each circular permutation, and an extra \order{\tilde{m}} factor arises from the fact that there are $m^2$ such permutations. The advantage here, though, is that we don't need to optimize masks having a very large size; we can do away with keeping $m$ a small constant because the scheme works for any $m$ such that $m \times m$ patches are sparse in some chosen dictionary. This scheme is, thus, more scalable in terms of the size of the input image. Therefore the effective dimension of the optimization problem in such a scheme is, in terms of the variables that matter, \order{T^2}.

It is worth mentioning that this simple idea has been largely ignored in literature concerning sensing matrix optimization. As mentioned in the introduction, previous attempts mostly use an average coherence minimization technique \cite{Duarte200907,Elad200610,Mordechay2014} for sensing matrices, and ignore not only the special structure of the matrices we are concerned with here but also the important issue of patch overlap. Moreover, none of these techniques are directly scalable for large images because they involve optimization problems in variables whose dimensions are at least of the order of image size. Sensing matrices can be designed at the patch level as well, for instance using information theoretic techniques as in~\cite{Carson2012,Renna2013,Weiss2007}, but the methods therein are not designed to account for either the special strcuture nor the issue of overlapping reconstruction. To the best of our knowledge, ours is the first piece of work to handle this important issue in a principled manner.

\section{\sc Validation and Results} \label{sec:results}
In this section, we show extensive results on reconstruction in the context of video compressive sensing and color image demosaicing. However, we start with testing the proposed framework visually. In each figure caption are values of the relative root mean square errors between the images and their reconstructed versions.

\subsection{Demonstrating a need for optimization}
We first show a need for optimizing codes by demonstrating source separation results on frames from a $301 \times 225$ 30 fps video scene. In this subsection, all reconstructions are performed in a non-overlapping way using Eq. \ref{eq:BP_patchwise}. 
%
%
With positive [0, 1] uniform random codes, the relative root mean square errors are around 8\% for each image. The results are shown in Fig.~\ref{fig:realUnif}.
\begin{figure}[!h]
\fourTwoAcross{pics/framework_visual/realUnif1in}{pics/framework_visual/realUnif2in}{pics/framework_visual/realUnif1est}{pics/framework_visual/realUnif2est}{1}
\caption{Real images, reconstructions with [0,1] uniform-random matrices. Up: input images, down: reconstructions. RRMSEs = 0.081 and 0.084 for the two reconstructions respectively. Compare with Fig.~\ref{fig:threeUnif}}
\label{fig:realUnif}
\end{figure}
Looking at these results, one notices that there is very little to no ghosting, that is, appearance of features from one image into the other, in the output images even when the images are very close to each other. This is a very desirable property in any algorithm that separates images from compressed video. These results are as good as the ones that Gaussian random codes generate.

\begin{figure}[!h]
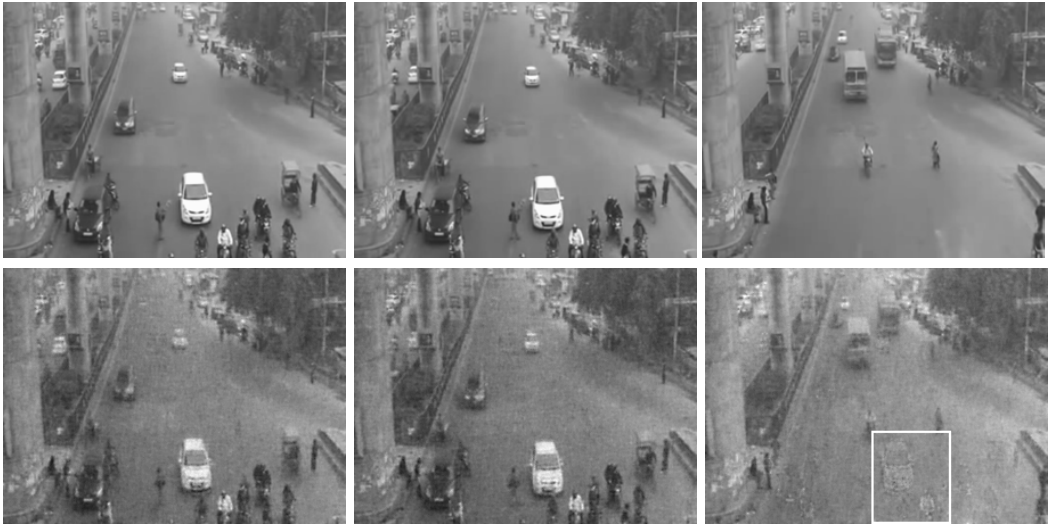

\sixThreeAcross{pics/framework_visual/threeImage1in}{pics/framework_visual/threeImage2in}{pics/framework_visual/threeImage3in}{pics/framework_visual/threeImage1est}{pics/framework_visual/threeImage2est}{pics/framework_visual/threeImage3est}{1.75}
\caption{Separating three images, uniform matrices. Up:~input images, down:~reconstructions. RRMSEs = 0.0814, 0.0926 and 0.0859 for the three reconstructions respectively. Compare with Fig.~\ref{fig:realUnif}. The ghost of the car from the first two frames appears in the white box in the reconstruction for the third frame.}
\label{fig:threeUnif}
\end{figure}

Next, we try separating three images from the same video with uniform matrices. The third frame contains some significant difference in content as compared to the earlier frames, and hence the reconstruction from data involving random codes reveals ghosting artifacts. For instance, where there is a car in the first two images, there is plain road in the third. See Fig.~\ref{fig:threeUnif}. Here, we notice ghosting artifacts appearing in the third frame (see the box in the third frame reconstruction). However, with better designed sensing matrices, one can think of getting rid of this effect, providing a motivation for matrix design in coded source separation. The relative root mean square errors here are worse, around 10\% for each image.


We perform a numerical comparison between our designed codes and random codes for various values of $s = \|\boldsymbol{X_i}\|_0/n$ and $T$. We randomly generate $T$ $s$-sparse (in 2D DCT) $8 \times 8$ signals $\{\boldsymbol{X_i}\}_{i=1}^{T}$, combine them using either random or designed codes to get $\boldsymbol{Y}$. Average relative root mean square errors on recovering the input signals from $\boldsymbol{Y}$ as a function of $s$ and $T$ are shown in Figs.~\ref{fig:frameworkNum} and \ref{fig:errorMapDesc}. Errors are near-zero in the region where both $T$ and $s$ are small, and one can expect reasonable quality reconstructions till $T = 4$ from random matrices. To increase $T$ further, we would need to optimize our sensing matrix appropriately, as is shown further in this paper.

\subsection{Coherence minimization}
\begin{figure}[!h]
\centering
\includegraphics[scale=0.25]{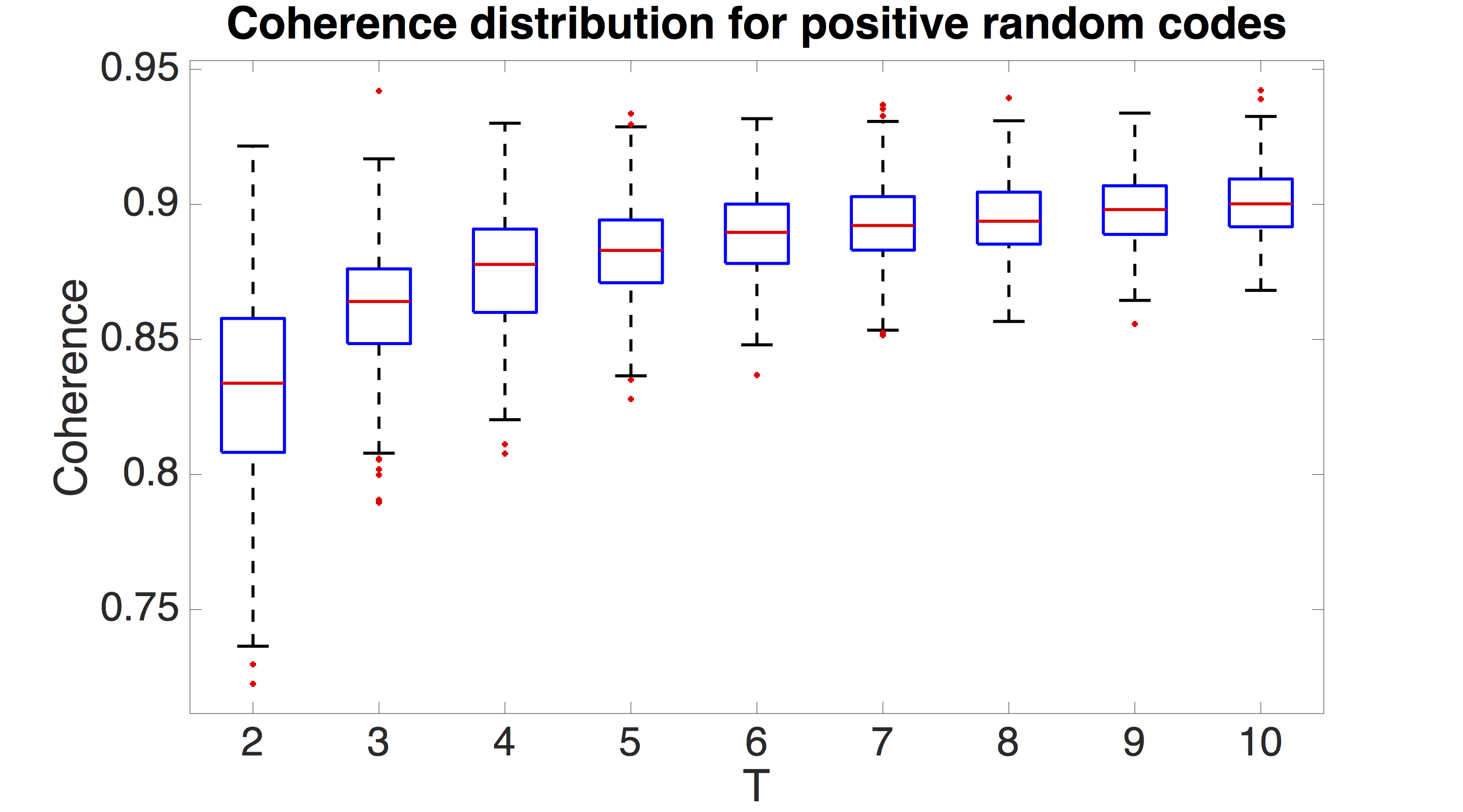}
\caption{Distribution of coherences for $8 \times 8$ random positive codes as a function of $T$}
\label{fig:randDistr}
\end{figure}

\begin{figure}[!h]
\centering
\includegraphics[scale=0.4]{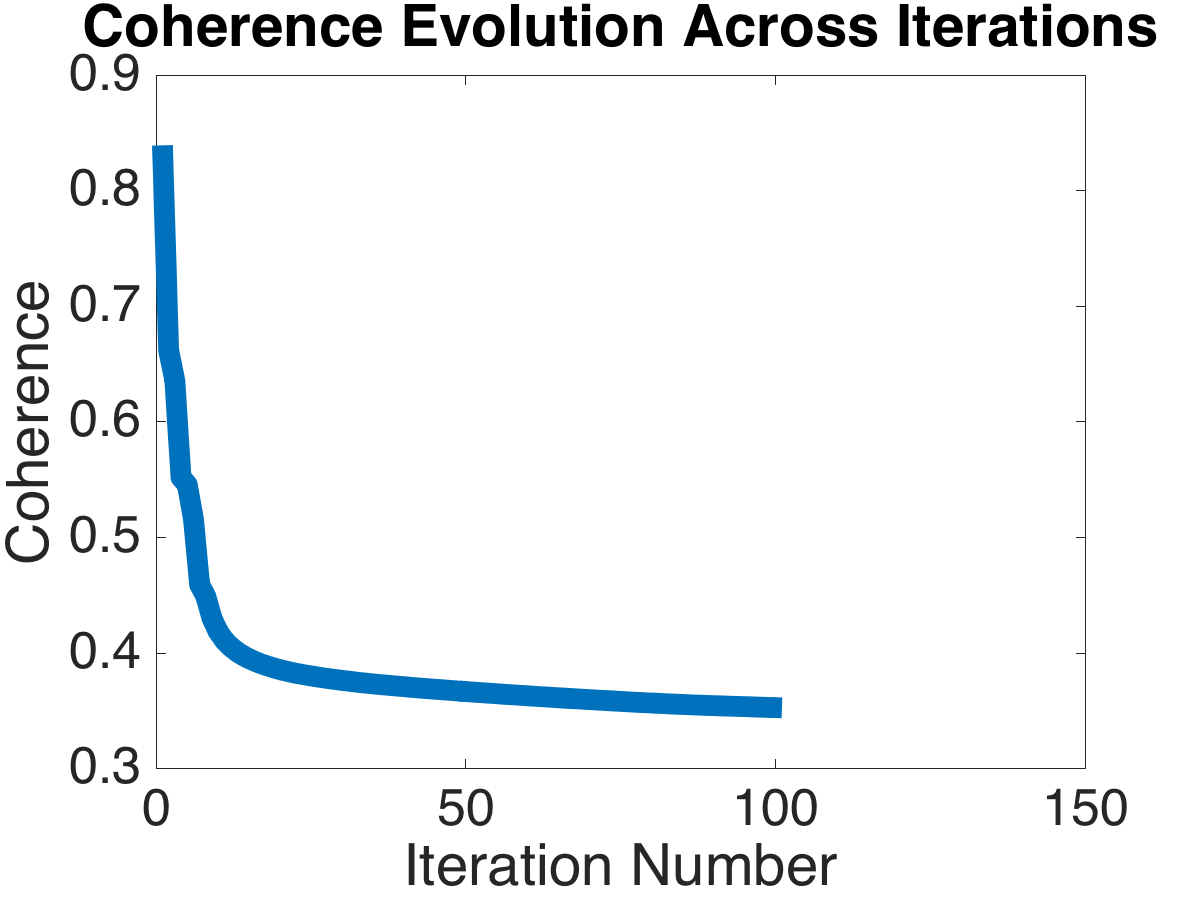}
\caption{Typical coherence decrease profile}
\label{fig:descPlot}
\end{figure}

The coherence of a uniform random matrix of the type we're interested in has a typical value around 0.8 for $8 \times 8$ codes. The distribution of these values is shown in the boxplot in Fig~\ref{fig:randDistr}.
The typical profile of descent on coherence from a random initialization is shown in Fig.~\ref{fig:descPlot}.

The minimum coherence we have been able to achieve in this scheme has been around 0.27 (for $T = 2$). It is interesting to note that all initialization instances lead to coherences (for $T = 2$) of at the most 0.35, and hence empirically yield nearly as good matrices.

\subsection{Circularly-symmetric coherence minimization}
\begin{figure}
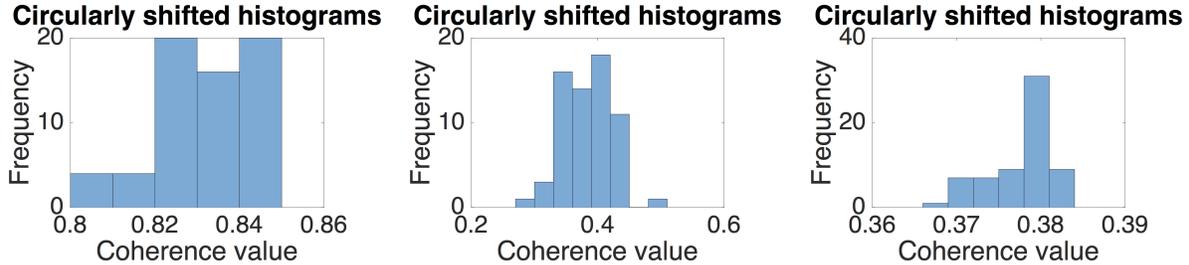

\centering
\threeAcross{pics/descent-circular/random-cohHist}{pics/descent-circular/noncircular-cohHist}{pics/descent-circular/circular-cohHist}{1}
\caption{Left to right: Circularly-shifted coherence histograms, in order, for random matrices, non-circularly optimized matrices and circularly optimized matrices}
\label{fig:cohHist}
\end{figure}

Again, we design $8 \times 8$ codes for $T = 2$. To show coherence improvement between positive random codes, and codes designed with and without circular permutations, we plot the distribution of coherences of $\Phi^{(\zeta)} D$ in Fig.~\ref{fig:cohHist} for all circular permutations $\zeta$. Note that even though the coherences of non-circularly designed matrices are much lower than positive random matrices, the maximum coherence among all permutations is quite large. The circularly-designed matrices, however, have permuted coherences clustered around a low value. We then expect good reconstruction with all circular permutations, yielding good expected reconstructions for images.

\subsection{Demosaicing}
To demonstrate the utility of this scheme, we show results on demosaicing RGB images. The general demosaicing problem involves addressing the difficulty that on a camera sensor, a single pixel can sense only one of the three R, G and B channels. Therefore, raw camera data needs to be interpolated to recover all the three channels. Traditional approaches to demosaicing involve the use of the Bayer pattern, which tiles a fixed [B, G; G, R] pattern over the image and use variants of algorithms like edge-directed interpolation which are tuned to the Bayer pattern. The Matlab \texttt{demosaic} function, for instance, uses \cite{Malvar2004}, which takes a gradient-corrected bilinear interpolated approach. However, recently a case has been made for panchromatic demosaicing~\cite{Hirakawa2008}, where we sense a linear combination of the three channels. In this paper, we use techniques from compressive recovery to perform demosaicing, i.e. to reconstruct the original color. However, it turns out that the Bayer pattern has very high mutual coherence, so it is unsuitable for compressive recovery. Here, we propose to design the mosaic patterns by minimizing mutual coherence.

\begin{figure}[!h]
\centering
\begin{tabular}{cc}
\includegraphics[scale=0.5]{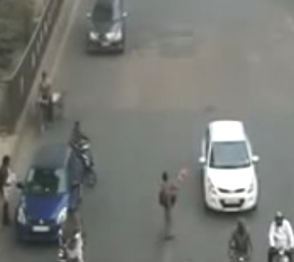}
\hspace{-10pt}
&
\includegraphics[scale=0.5]{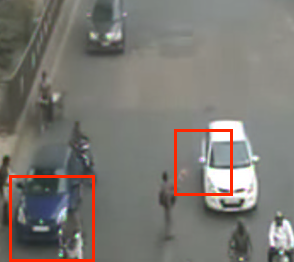}
\hspace{-10pt}
\\
\includegraphics[scale=0.5]{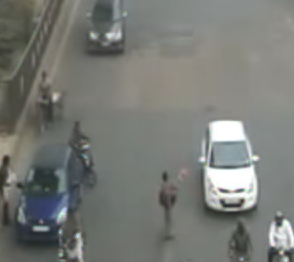}
\hspace{-10pt}
&
\includegraphics[scale=0.5]{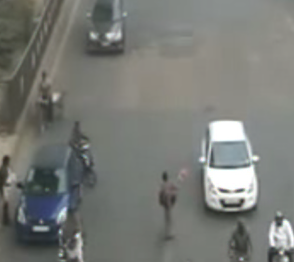}
\hspace{-10pt}
\\
\begin{tabular}{cc}
\includegraphics[scale=0.5]{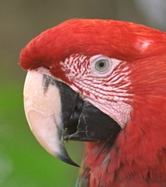}
\hspace{-10pt}
&
\includegraphics[scale=0.5]{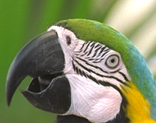}
\hspace{-10pt}
\end{tabular}
&
\begin{tabular}{cc}
\includegraphics[scale=0.5]{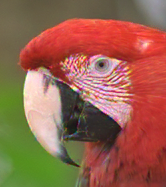}
\hspace{-10pt}
&
\includegraphics[scale=0.5]{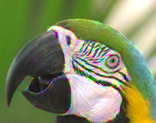}
\hspace{-10pt}
\end{tabular}
\\
\begin{tabular}{cc}
\includegraphics[scale=0.5]{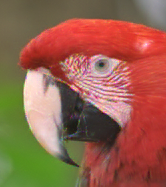}
\hspace{-10pt}
&
\includegraphics[scale=0.5]{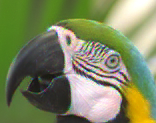}
\hspace{-10pt}
\end{tabular}
&
\begin{tabular}{cc}
\includegraphics[scale=0.5]{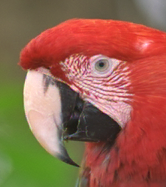}
\hspace{-10pt}
&
\includegraphics[scale=0.5]{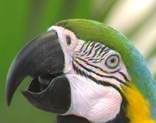}
\hspace{-10pt}
\end{tabular}
\end{tabular}
\caption{Demosaicing closeups: In each group of images - original image (top left), reconstructions with random matrices (top right), with circularly optimized matrices (bottom right), and with non-circularly optimized matrices (bottom left).}
\label{fig:smallScaleDemos}
\end{figure}

We design $8 \times 8$ codes for linearly combining the three channels using our method and visually compare overlapping reconstructions. 
\begin{figure}[!h]
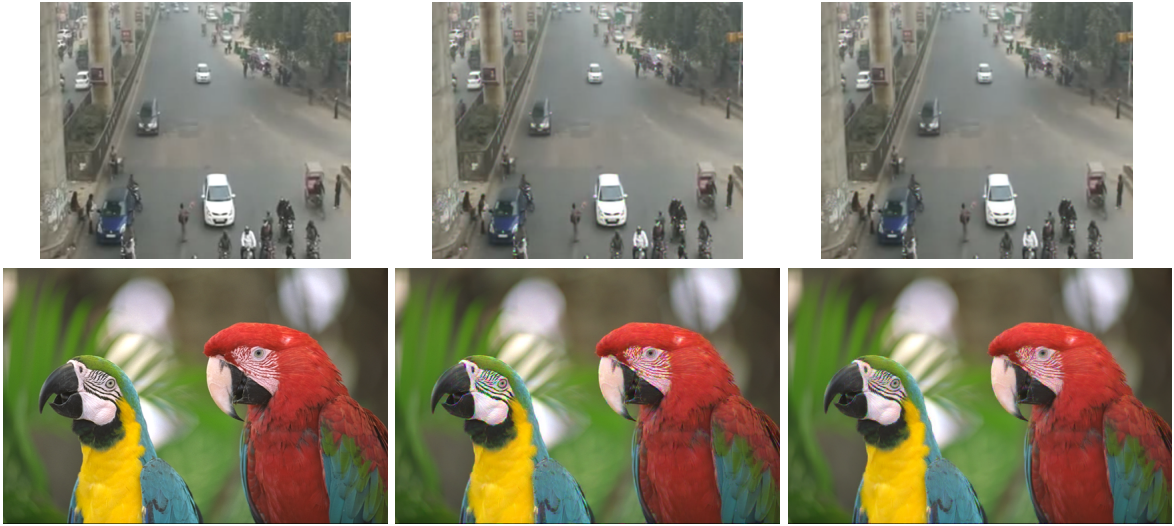

\sixThreeAcross{pics/demosaicing/736503_2161-DemosaicDesignedOvp-in}{pics/demosaicing/736503_2926-DemosiacRandomOvp-out}{pics/demosaicing/736503_2161-DemosaicDesignedOvp-out}{pics/demosaicing/736503_4279-ParrotDemosaicDesignedOvp-in}{pics/demosaicing/736503_4782-ParrotDemosiacRandomOvp-out}{pics/demosaicing/736503_4279-ParrotDemosaicDesignedOvp-out}{1.75}
\caption{Demosaicing. Left: original images, middle: reconstructions with random matrices, right: reconstructions with non-circularly designed matrices. RRMSEs = 0.0189 and 0.0189 for upper image with random matrices and non-circularly designed matrices respectively. RRMSEs = 0.0517 and 0.0510 for lower image with random matrices and non-circularly designed matrices respectively}
\label{fig:demos}
\end{figure}
As Figs.~\ref{fig:demos} and \ref{fig:smallScaleDemos} show, results from the designed case are more faithful to the ground-truth than the random reconstructions are. The random reconstructions show (more) color artifacts, especially in areas where the input image varies a lot (car headlights in the top image, around parrot eyes in the bottom). Our designed codes do not show as many color artifacts. The relative root mean square errors don't differ much for these two cases, but subtle details of color are better preserved by our matrices. (Indeed, it is well-known in the image processing community that object quality metrics such as RMSE can differ from human perception). In Fig.~\ref{fig:smallScaleDemos}, notice in the first case the green artifacts near car headlights and the leftmost cyclist in the results with random matrices, while that area is better reconstructed with our designed matrices. The car headlight area on the car at the right is also better reconstructed by our matrices. In the bottom, notice less color artifacts in the densely-varying area near the eye and on the bottom part of the beak.

\subsection{Video compressed sensing}
\begin{figure}[!h]
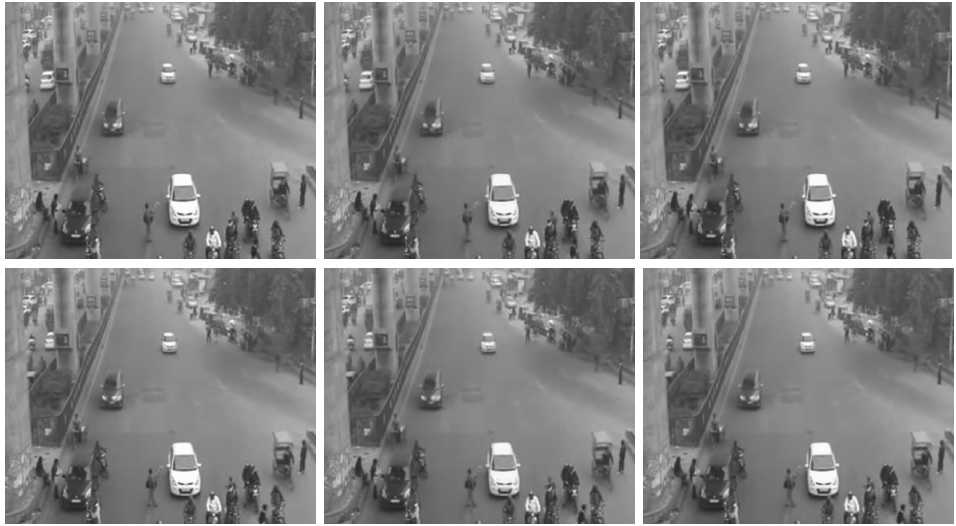

\sixThreeAcross{pics/descent/736497_3251-RandomOvp-2-1-in}{pics/descent/736497_3251-RandomOvp-2-1-out}{pics/descent/736496_9893-DesignedOvp-2-1-out}{pics/descent/736497_3251-RandomOvp-2-2-in}{pics/descent/736497_3251-RandomOvp-2-2-out}{pics/descent/736496_9893-DesignedOvp-2-2-out}{1.75}
\caption{Optimized output from combining two close images. Left: inputs, middle, right: reconstructions, in order, with random matrices and non-circularly optimized matrices. RRMSEs = 0.0277, 0.0139 for frame 1 and 0.0280, 0.0136 for frame 2 for random matrices and non-circularly optimized matrices respectively}
\label{fig:optTwo}
\end{figure}

\begin{figure}[]
\setlength{\tmpLength}{0.98\textwidth}
\setlength{\tmpLength}{0.12\tmpLength}
\def\arraystretch{1}
\setlength\tabcolsep{2pt}
\centerline{
\begin{tabular}{ccc}
\includegraphics[height=1.7\tmpLength]{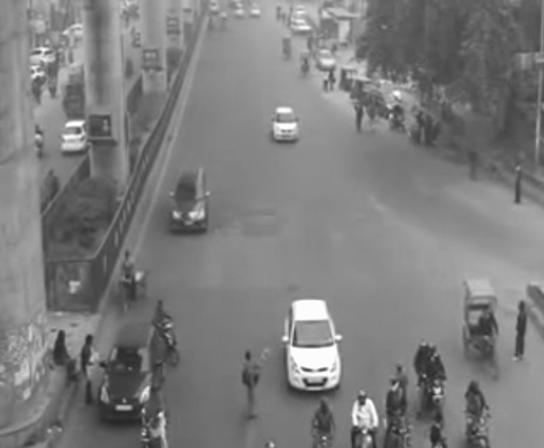} &
\includegraphics[height=1.7\tmpLength]{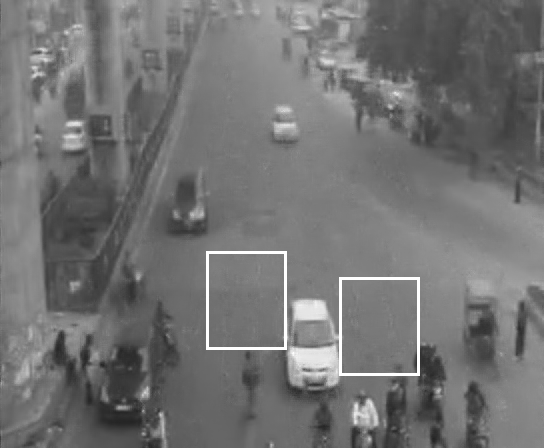} &
\includegraphics[height=1.7\tmpLength]{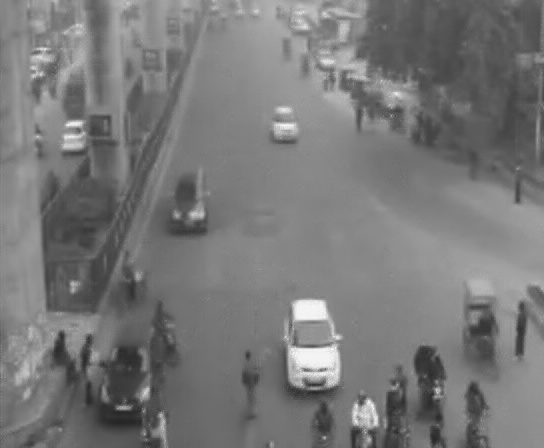} \\
\includegraphics[height=1.7\tmpLength]{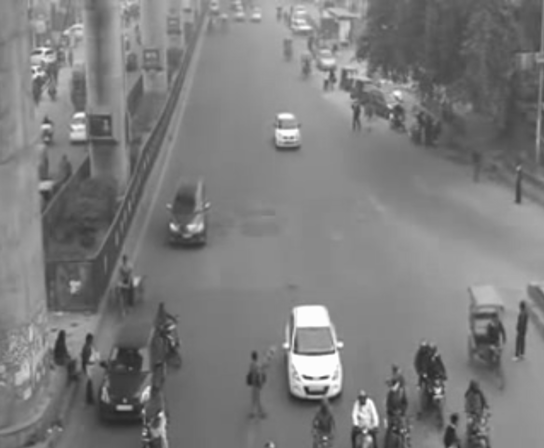} &
\includegraphics[height=1.7\tmpLength]{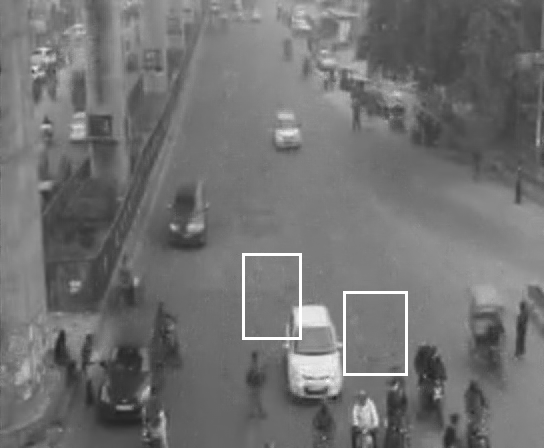} &
\includegraphics[height=1.7\tmpLength]{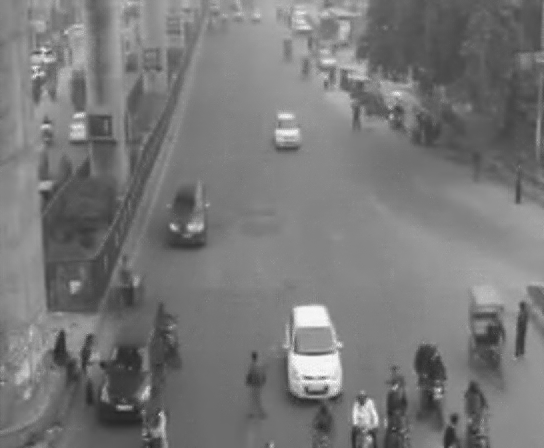} \\
\includegraphics[height=1.7\tmpLength]{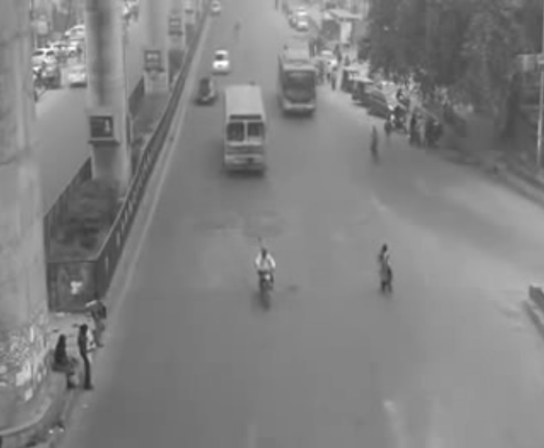} &
\includegraphics[height=1.7\tmpLength]{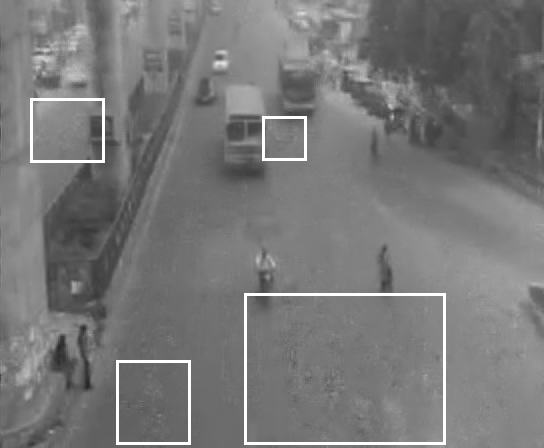} &
\includegraphics[height=1.7\tmpLength]{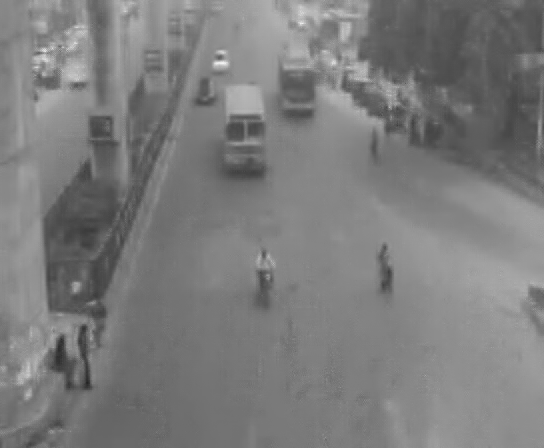} \\
\includegraphics[height=1.7\tmpLength]{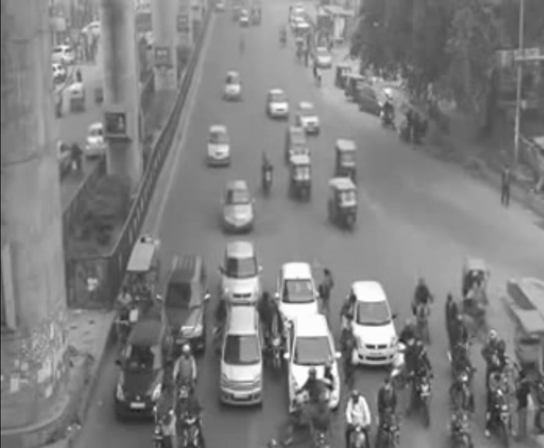} &
\includegraphics[height=1.7\tmpLength]{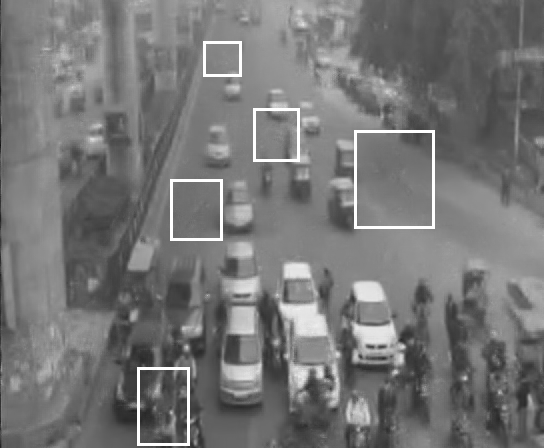} &
\includegraphics[height=1.7\tmpLength]{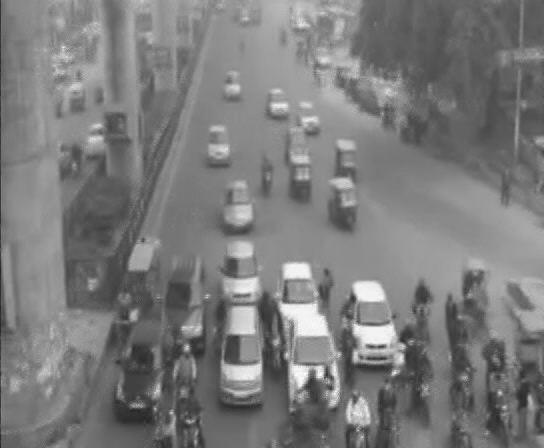} \\
\includegraphics[height=1.7\tmpLength]{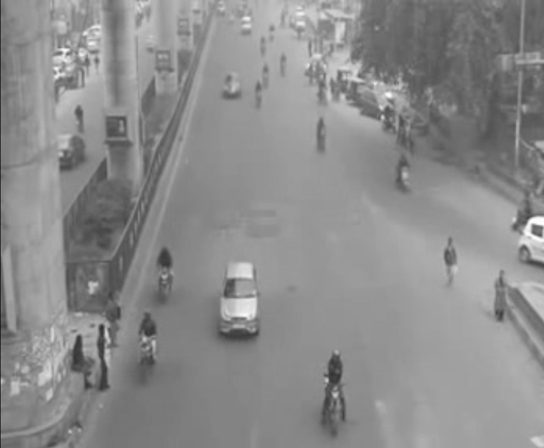} &
\includegraphics[height=1.7\tmpLength]{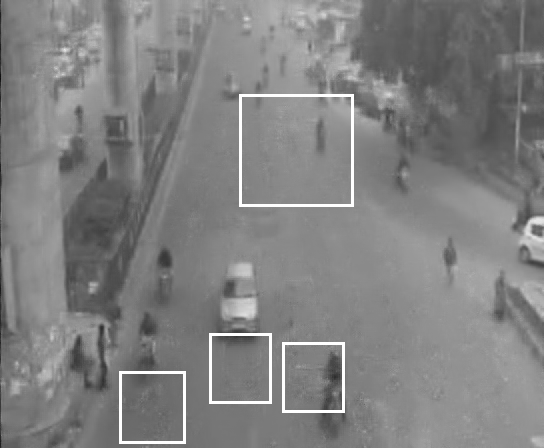} &
\includegraphics[height=1.7\tmpLength]{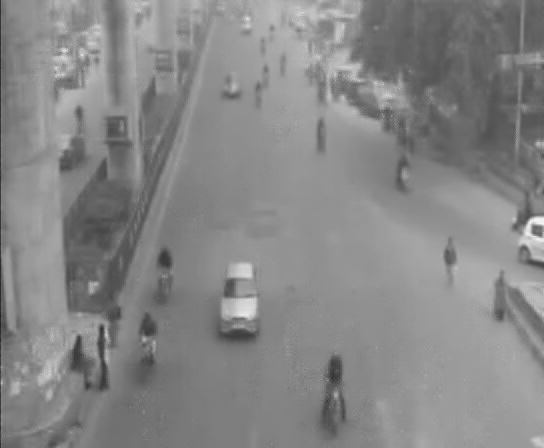} \\
\includegraphics[height=1.7\tmpLength]{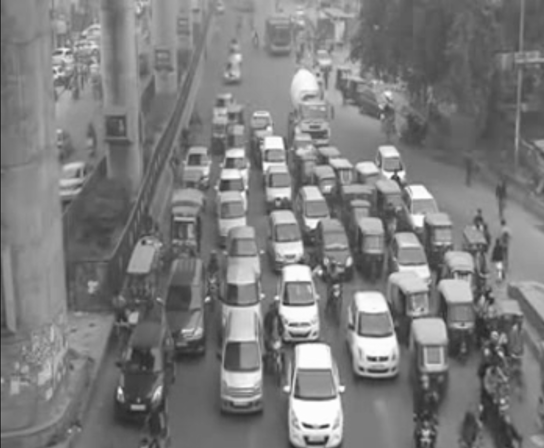} &
\includegraphics[height=1.7\tmpLength]{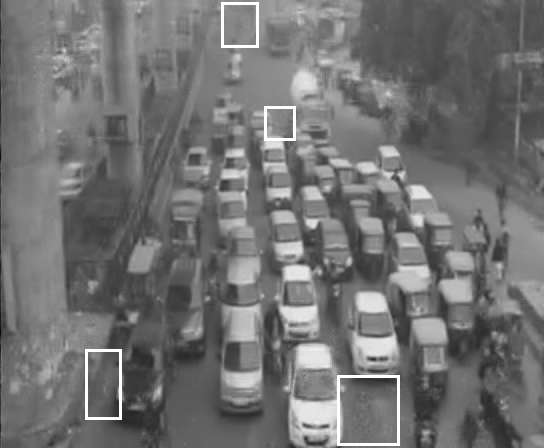} &
\includegraphics[height=1.7\tmpLength]{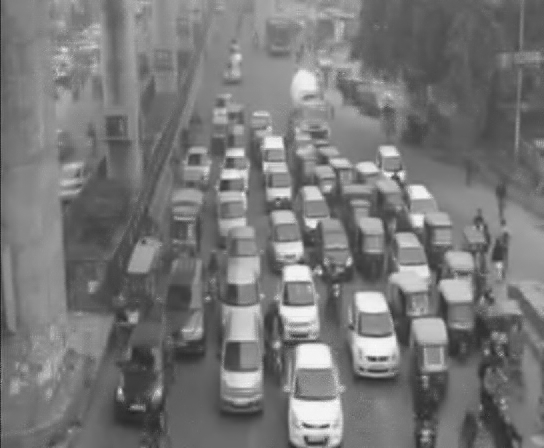} 
\end{tabular}
} 
\caption{Optimized output from combining six not necessarily close images. Left: inputs, middle: reconstructions with random matrices, right: reconstructions with non-circularly optimized matrices. Respective RRMSES = 0.1053 and 0.0574 for frame 1, 0.1051 and 0.0532 for frame 2, 0.1016 and 0.0558 for frame 3, 0.1090 and 0.607 for frame 4, 0.1012 and 0.0496 for frame 5, 0.1100 and 0.0650 for frame 6 for random matrices and non-circularly optimized matrices.}
\label{fig:optSix}
\end{figure}

We now visually validate that our designed matrices perform better than positive random matrices. We design $8 \times 8$ codes and tile them, reconstructing patchwise with overlapping patches. An example of this running on six not necessarily close frames in a video is shown in Fig.~\ref{fig:optSix} (Fig.~\ref{fig:optTwo} shows an example for $T = 2$). This is a challenging test case involving large object motion or significant structural changes in the scene. Ghosting artifacts marked out in Fig.~\ref{fig:optSix} in white boxes in the random matrix reconstructions are absent or lower in the designed matrix reconstructions. These outputs show that on the large scale, we do as well as random matrices for low $T$ and better for high $T$. For a small scale comparison, see Figs.~\ref{fig:smallScaleComp1} and \ref{fig:smallScaleComp2}.

\begin{figure}[!h]
\centering
\includegraphics[scale=0.25]{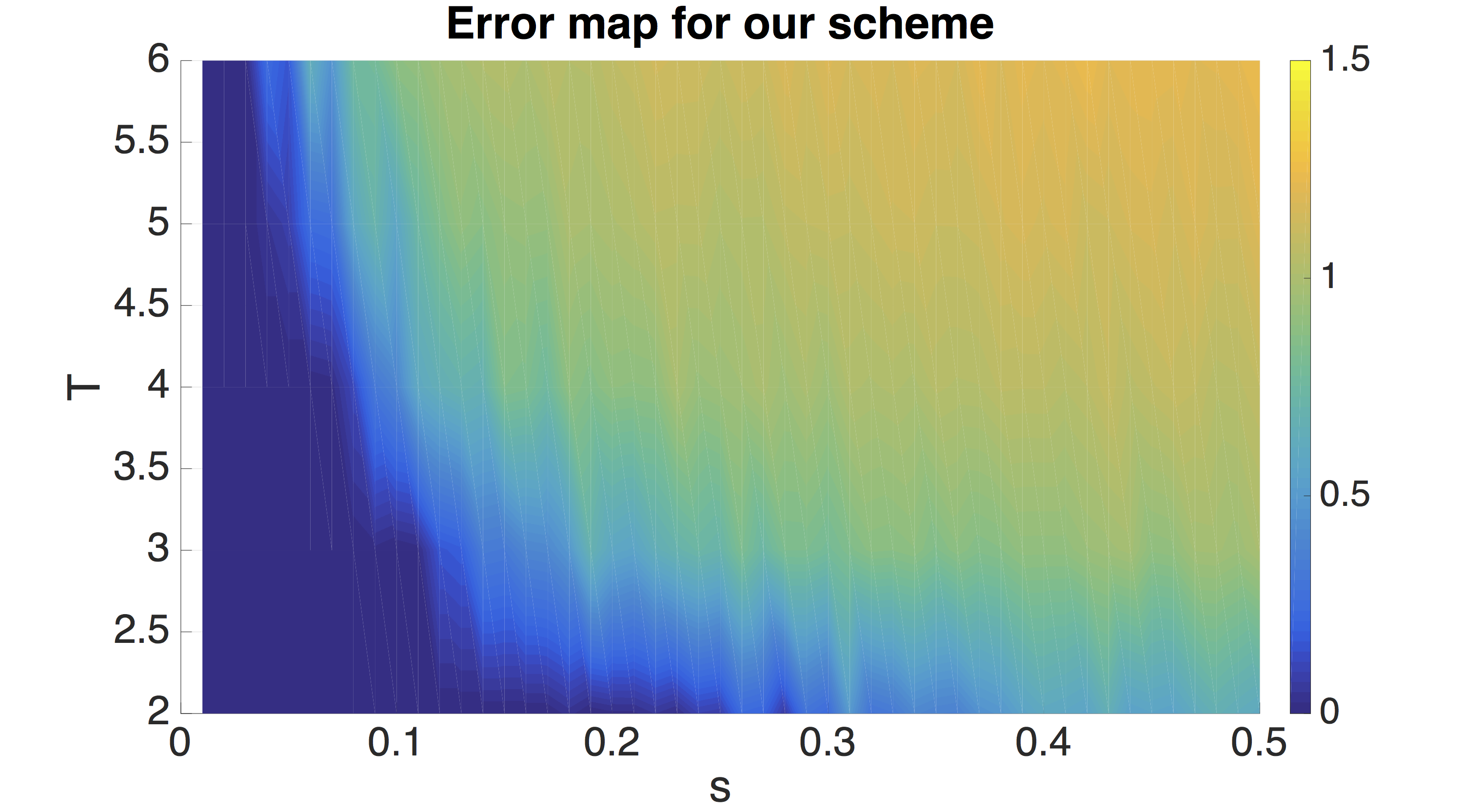}
\caption{Average relative root mean square errors in our scheme as a function of $s$ and $T$ with positive random matrices. Compare with Figures \ref{fig:errorMapDesc} and \ref{fig:diffMapDesc}.}
\label{fig:frameworkNum}
\end{figure}

\begin{figure}[!h]
\centering
\includegraphics[scale=0.25]{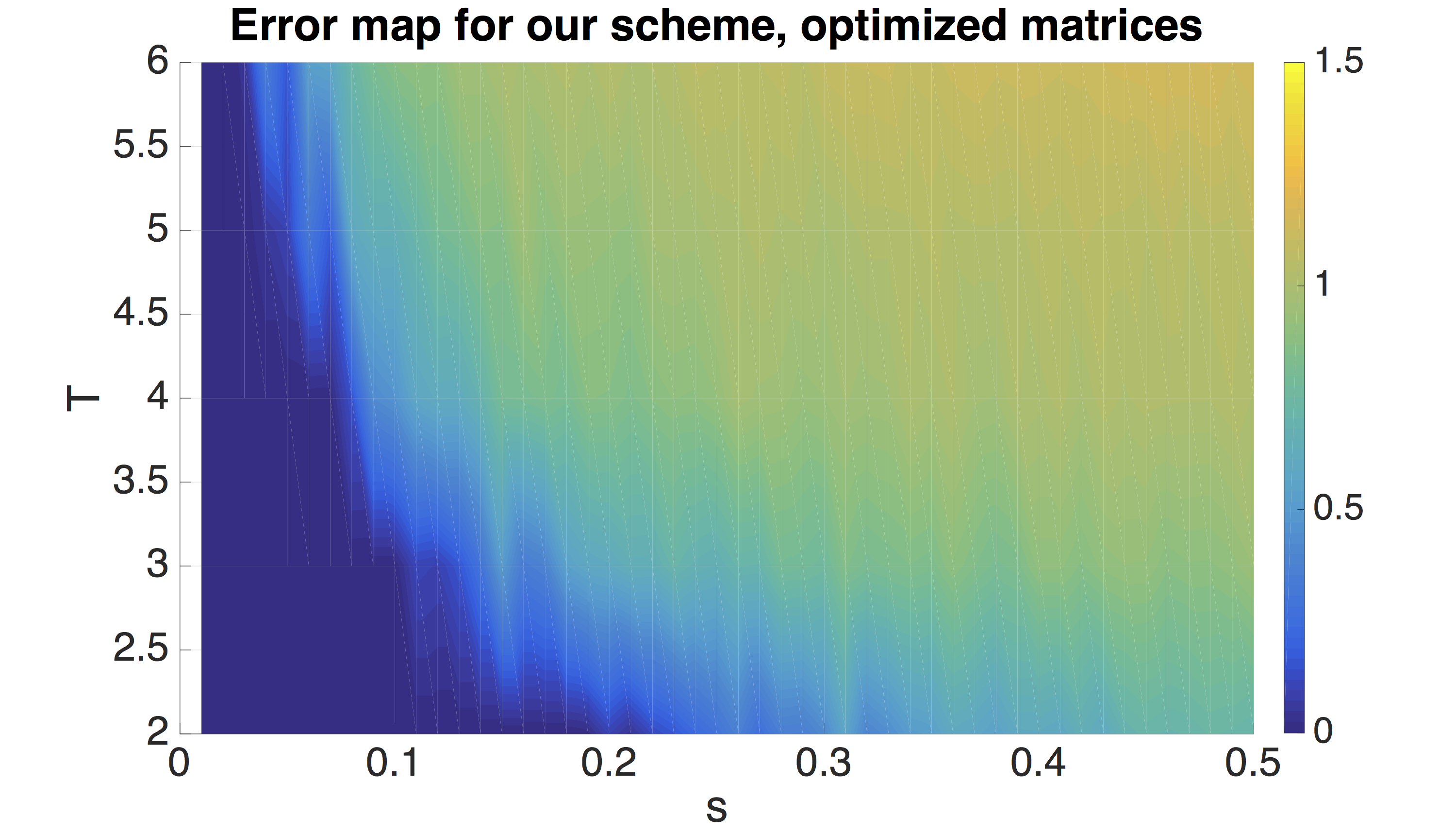}
\caption{Error map for optimized codes as a function of $s$ and $T$. Compare with Figures \ref{fig:frameworkNum} and \ref{fig:diffMapDesc}.}
\label{fig:errorMapDesc}
\end{figure}

\begin{figure}[!h]
\centering
\includegraphics[scale=0.2]{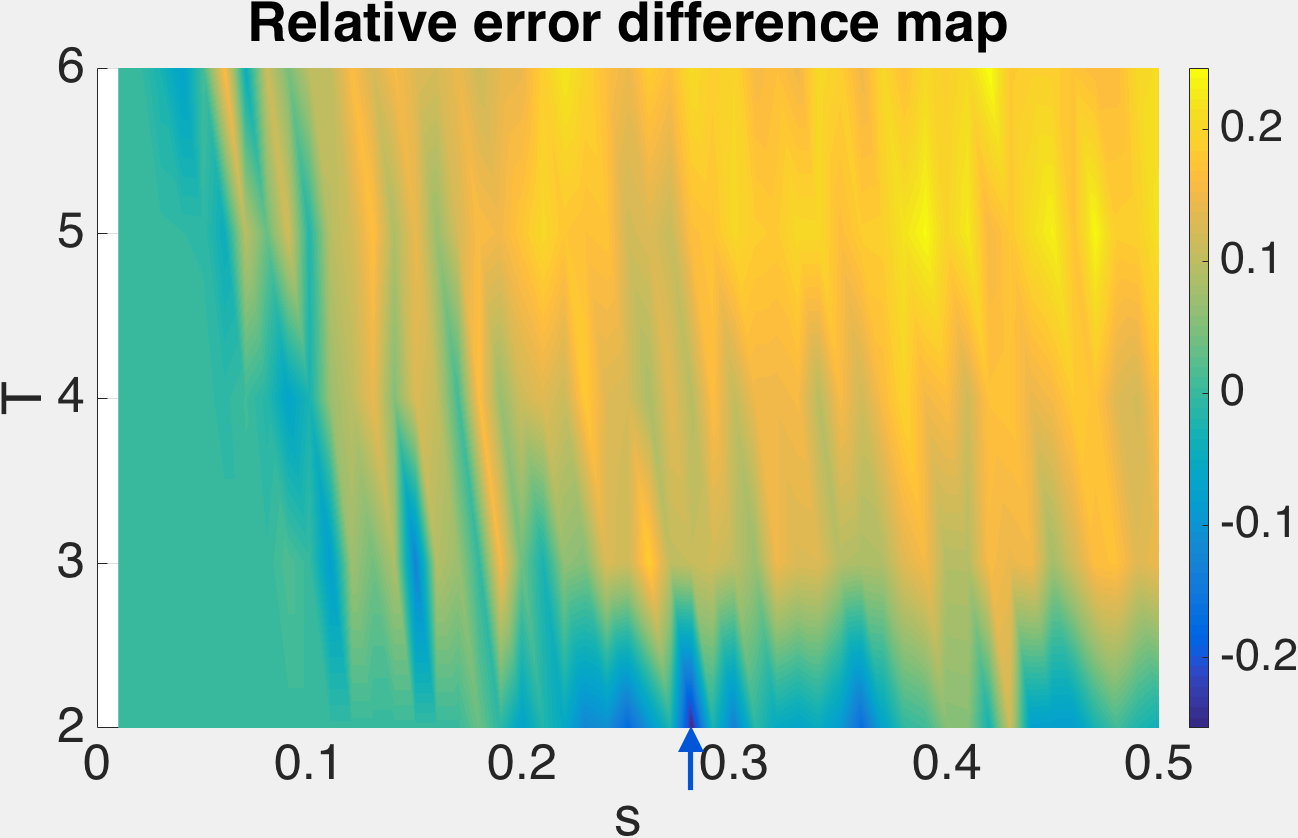}
\caption{Average RRMSE using random codes minus average RRMSE using optimized codes, as a function of $s$ and $T$. Notice that the designed matrices outperform random matrices in most places except the blue regions, where both types of matrices yield low RMSE.}
\label{fig:diffMapDesc}
\end{figure}

Finally, we perform a numerical comparison similar to the one in Fig.~\ref{fig:frameworkNum}. The resulting error map is shown in Fig.~\ref{fig:errorMapDesc}. On an average, we see that we perform better than the random case at high $T$ and $s$, which is the region we aim to optimize for. For a comparison, the difference between the two errors is shown in Fig.~\ref{fig:diffMapDesc}. The RRMSE difference does produce significant changes in subtle texture as seen in Figs.~\ref{fig:smallScaleComp1} and \ref{fig:smallScaleComp2}. In the deep blue low $T$ regions in Fig.~\ref{fig:diffMapDesc}, for instance near $s=0.3$ (see blue arrow), where random matrices seem to outperform us, the reconstruction errors are small enough for both random and designed matrices to cause no significant depreciation in our reconstruction quality in comparison to random.

\begin{figure}[!h]
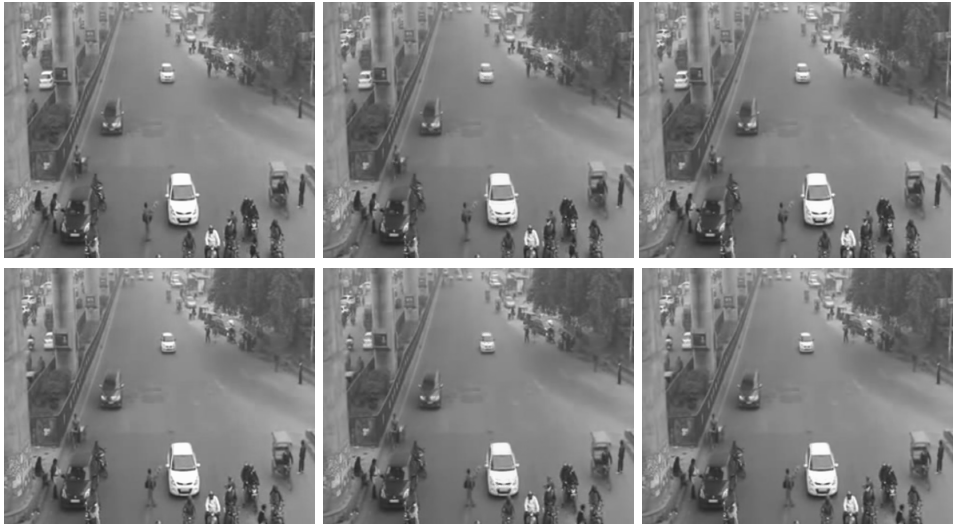

\sixThreeAcross{pics/descent/736497_3251-RandomOvp-2-1-in}{pics/descent-circular/736497_3251-RandomOvp-2-1-out}{pics/descent-circular/736509_9723-DesignedOvp-2-1-out}{pics/descent/736497_3251-RandomOvp-2-2-in}{pics/descent-circular/736497_3251-RandomOvp-2-2-out}{pics/descent-circular/736509_9723-DesignedOvp-2-2-out}{1.75}
\caption{Circularly optimized output from combining two close images. Left to right: reconstructions, in order, with random matrices and circularly optimized matrices. RRMSEs = 0.0277, 0.0139 for frame 1; and 0.0280, 0.0136 for frame 2 for random matrices and circularly optimized matrices respectively}
\label{fig:optCircTwo}
\end{figure}

Similar to the above analysis for non-circularly designed matrices, we validate our circularly designed matrices visually. Following the same conventions, here is an output for the $T=2$ case [Fig.~\ref{fig:optCircTwo}].

We now look at reconstructions from random and both classes of our designed matrices on a small scale. As a first example, we show a close-up from the car video sequence shown earlier [Fig.~\ref{fig:smallScaleComp1}]. Note, to start off, that the reconstruction of the numberplate and headlight area is much clearer in our case than the random matrix case. Further, notice the presence of major ghosting in the random case, especially near the rear-view mirrors, bonnet (marked by arrows) and headlights (marked by boxes), while our reconstructions remain free of these artifacts. Adding circular optimization to the picture further improves image quality especially in the bonnet area, where the non-circular reconstruction is slightly splotchy. Next, in Fig.~\ref{fig:smallScaleComp2}, which is a smaller part of the same image, the superiority of our reconstruction is clearer, with the circular optimization smoothing out blotchier parts of the bonnet.
\begin{figure}[!h]
\twoHeight{pics/comparison/comp1-1}{pics/comparison/comp1-2}{80pt}
\caption{Close-ups showing subtle texture preservation with optimized matrices, example 1. Left to right: inputs, reconstructions, in order, with random matrices, non-circularly optimized matrices and circularly optimized matrices}
\label{fig:smallScaleComp1}
\end{figure}
\begin{figure}[!h]
\twoHeight{pics/comparison/comp2-1}{pics/comparison/comp2-2}{80pt}
\caption{Close-ups showing subtle texture preservation with optimized matrices, example 2. Left to right: inputs, reconstructions, in order, with random matrices, non-circularly optimized matrices and circularly optimized matrices}
\label{fig:smallScaleComp2}
\end{figure}

\subsection{Comparison with Other Techniques}
We emphasize that a direct comparison with earlier methods \cite{Elad200610,Duarte200907,Li2017,Abolghasemi2012,Hong2016} is not possible as they do not tackle the same computational problem that we deal with in this paper - namely these earlier works do not impose the structure that we impose on the designed matrices. In fact, the aforementioned techniques cannot be modified to handle the special structure of the matrices we deal with. For example, \cite{Duarte200907,Hong2016} provides a closed form expression for the sensing matrix in the form of an SVD, thereby making it difficult to impose additional constraints.

\subsection{Reproducible Research}
Most code used in generating results and optimizing sensing matrices in this paper lives in the Bitbucket repository at \href{https://bitbucket.org/alankarkotwal/coded-sourcesep}{\texttt{alankarkotwal/coded-sourcesep}}~\cite{Implement}. Gradient descent lives in the \texttt{src/descent} folder, circularly-symmetric gradient descent in \texttt{src/descent-circular} and reconstruction code in \texttt{src/circular}.

\section{\sc Conclusion} \label{sec:concl}
We cast the video compressed sensing or color image demosaicing problem as one of separation of coded linear combinations of signals sparse in a given basis. We evaluated this scheme and found it works well for low sparsity levels, and yields reasonably visually good reconstructions. However, especially at high $T$, we found that random matrices aren't good enough, motivating design of sensing matrices based on optimizing mutual coherence, a well-known measure of matrix `quality' in compressive recovery.

We then provided an analytical expression for the coherence of the sensing matrix in the coded source separation scheme and optimized for coherence using these. Results showed better quality and less ghosting visually, and less error numerically. However as image size increased, the optimization problem became rapidly intractable, so we settled for optimizing small masks such that they have small coherence in all circular permutations, so they can be tiled for overlapping patch-wise reconstruction. We highlighted the importance of these circular permutations for improving reconstruction performance. 

Future work will involve matrix design using criteria that make explicit use of specific image priors as well as likelihood functions that reflect specific noise models occuring in imaging applications. Our method can also be extended to handle multi-spectral image demosaicing or reconstruction. 

\appendix
\section{Derivation of coherence expression} \label{App:derCoh}
Recalling our definitions, we call the index varying from $1$ to $T$ as $\mu$ or $\nu$, and the index varying from $1$ to $n$ as $\alpha$, $\beta$ or $\gamma$. The $\mu^\text{th}$ block of $\boldsymbol{\Phi}$ is thus $\boldsymbol{\Phi_\mu}$. Let the $\beta^\text{th}$ diagonal element of $\boldsymbol{\Phi_\mu}$ be $\phi_{\mu\beta}$. Define the $\alpha^\text{th}$ column of $\boldsymbol{D^T}$ to be $\boldsymbol{d_\alpha}$. Thus, the Gram matrix $\boldsymbol{\tilde{M}} = \boldsymbol{\Psi^T \Phi^T \Phi \Psi}$ has the block structure
\begin{align*}
\boldsymbol{\tilde{M}_{\mu \nu}} &= \boldsymbol{D^T \Phi_\mu^T \Phi_\nu D} \\
&= \boldsymbol{D^T \Phi_\mu \Phi_\nu D} \\
&= \begin{pmatrix}
\boldsymbol{d_1} & \boldsymbol{d_2} & \hdots & \boldsymbol{d_n}
\end{pmatrix}
\begin{pmatrix}
\phi_{\mu 1} \phi_{\nu 1} & 0 & \hdots & 0 \\
0 & \phi_{\mu 2} \phi_{\nu 2} & \hdots & 0 \\
\vdots & \vdots & \ddots & \vdots \\
0 & 0 & \hdots & \phi_{\mu n} \phi_{\nu n}
\end{pmatrix}
\begin{pmatrix}
\boldsymbol{d_1^T} \\
\boldsymbol{d_2^T} \\
\vdots \\
\boldsymbol{d_n^T}
\end{pmatrix} \\
&= \begin{pmatrix}
\boldsymbol{d_1} & \boldsymbol{d_2} & \hdots & \boldsymbol{d_n}
\end{pmatrix}
\begin{pmatrix}
\phi_{\mu 1} \phi_{\nu 1} \boldsymbol{d_1^T} \\
\phi_{\mu 2} \phi_{\nu 2} \boldsymbol{d_2^T} \\
\vdots \\
\phi_{\mu n} \phi_{\nu n} \boldsymbol{d_n^T}
\end{pmatrix} \\
&= \sum_{\alpha = 1}^{n} \phi_{\mu \alpha} \phi_{\nu \alpha} \boldsymbol{d_\alpha d_\alpha^T}
\end{align*}

\noindent The $\beta\gamma^\text{th}$ element of $\boldsymbol{\tilde{M}_{\mu \nu}}$, thus, is (defining $d_\alpha(\beta)$ as the $\beta^\text{th}$ element of $\boldsymbol{d_\alpha}$)
\begin{align}
\tilde{M}_{\mu \nu}(\beta\gamma) &= \sum_{\alpha = 1}^{n} \phi_{\mu \alpha} \phi_{\nu \alpha} d_\alpha (\beta) d_\alpha (\gamma)
\end{align}

\noindent Now we need to normalize the columns of $\boldsymbol{\Phi \Psi}$. Squared column norms are diagonal elements of $\boldsymbol{\tilde{M}_{\mu \nu}}$. So the product of the squared norms of the $\beta^\text{th}$ column of the $\mu^\text{th}$ block and the $\gamma^\text{th}$ column of the $\nu^\text{th}$ block is (call this $\xi_{\mu \nu}^2 (\beta \gamma)$)
\begin{align}
\xi_{\mu \nu}^2 (\beta \gamma) &= \left( \sum_{\alpha = 1}^{n} \phi_{\mu \alpha}^2 d^2_\alpha (\beta) \right) \left( \sum_{\tau = 1}^{n} \phi_{\nu \tau}^2 d^2_\tau (\gamma) \right)
\end{align}

\noindent Let the normalized Gram matrix be $\boldsymbol{M}$. Thus, following the same conventions as above (define the numerator of the expression to be $\chi_{\mu \nu} (\beta \gamma)$), 
\begin{align}
M_{\mu \nu}(\beta\gamma) &= \frac{\sum_{\alpha = 1}^{n} \phi_{\mu \alpha} \phi_{\nu \alpha} d_\alpha (\beta) d_\alpha (\gamma)}{\sqrt{\left( \sum_{\alpha = 1}^{n} \phi_{\mu \alpha}^2 d^2_\alpha (\beta) \right) \left( \sum_{\tau = 1}^{n} \phi_{\nu \tau}^2 d^2_\tau (\gamma) \right)}} = \frac{\chi_{\mu \nu} (\beta \gamma)}{\xi_{\mu \nu} (\beta \gamma)}
\end{align}

\noindent Finally, using the square soft-max function to deal with the \texttt{max} in the coherence expression, we get the squared soft coherence $\mathcal{C}$ to be
\begin{align}
\mathcal{C} = \frac{1}{\theta} \log \left[ \sum_{\mu=1}^{T} \sum_{\nu=1}^{\mu-1} \sum_{\beta=1}^{n} \sum_{\gamma=1}^{n} e^{\theta M_{\mu \nu}^2(\beta\gamma)} + \sum_{\mu=1}^{T} \sum_{\beta=1}^{n} \sum_{\gamma=1}^{\beta - 1} e^{\theta M_{\mu \mu}^2(\beta\gamma)} \right]
\end{align}

\noindent In the above, the first term corresponds to all ($\mu > \nu$) blocks that are `below' the block diagonal. Here, we consider all terms in the given block for the maximum. The second term corresponds to ($\mu = \nu$) blocks on the block diagonal. Here, we consider only consider ($\beta > \gamma$) below-diagonal elements for the maximum.

\section{Derivation of coherence derivatives} \label{App:derCohDer}
Differentiating the expression for the squared soft coherence above, we get
\begin{equation}
\begin{split}
\frac{d\mathcal{C}(\Phi)}{d\phi_{\delta \epsilon}} = \frac{1}{\theta e^{\theta \mathcal{C}(\Phi)}} \left[ \sum_{\mu=1}^{T} \sum_{\nu=1}^{\mu-1} \sum_{\beta=1}^{n} \sum_{\gamma=1}^{n} 2\theta e^{\theta M_{\mu \nu}^2(\beta\gamma)} M_{\mu \nu}(\beta\gamma) \frac{dM_{\mu \nu}(\beta\gamma)}{d\phi_{\delta \epsilon}} \right. \\
\left. + \sum_{\mu=1}^{T} \sum_{\beta=1}^{n} \sum_{\gamma=1}^{\beta - 1} 2\theta e^{\theta M_{\mu \mu}^2(\beta\gamma)} M_{\mu \mu}(\beta\gamma) \frac{\theta M_{\mu \mu}(\beta\gamma)}{d\phi_{\delta \epsilon}} \right]
\end{split}
\end{equation}

\noindent Next, we calculate the derivatives in the above equation, ${dM_{\mu \nu}(\beta\gamma)}/{d\phi_{\delta \epsilon}}$. Define the numerator of the expression for $M_{\mu \nu}(\beta\gamma)$ as $\chi_{\mu \nu}(\beta\gamma)$, and thus, $M_{\mu \nu}(\beta\gamma) = \chi_{\mu \nu}(\beta\gamma)/\xi_{\mu \nu}(\beta\gamma)$. Clearly,
\begin{align}
\frac{dM_{\mu \nu}(\beta\gamma)}{d\phi_{\delta \epsilon}} = \frac{\xi_{\mu \nu}(\beta\gamma) \frac{d\chi_{\mu \nu}(\beta\gamma)}{d\phi_{\delta \epsilon}} - \chi_{\mu \nu}(\beta\gamma)\frac{d\xi_{\mu \nu}(\beta\gamma)}{d\phi_{\delta \epsilon}}}{\xi_{\mu \nu}(\beta\gamma)^2}
\end{align}

\noindent Next,
\begin{align*}
\frac{d\chi_{\mu \nu}(\beta\gamma)}{d\phi_{\delta \epsilon}} &= \frac{d}{d\phi_{\delta \epsilon}} \sum_{\alpha = 1}^{n} \phi_{\mu \alpha} \phi_{\nu \alpha} d_\alpha (\beta) d_\alpha (\gamma) \\
&= \sum_{\alpha = 1}^{n} d_\alpha (\beta) d_\alpha (\gamma) \frac{d}{d\phi_{\delta \epsilon}} \left( \phi_{\mu \alpha} \phi_{\nu \alpha} \right)
\end{align*}

\noindent Notice that a term in the above summation can be non-zero only if $\alpha = \epsilon$. Thus,
\begin{align*}
\frac{d\chi_{\mu \nu}(\beta\gamma)}{d\phi_{\delta \epsilon}} &= d_\epsilon (\beta) d_\epsilon (\gamma) \frac{d}{d\phi_{\delta \epsilon}} \left( \phi_{\mu \epsilon} \phi_{\nu \epsilon} \right) \\
&= d_\epsilon (\beta) d_\epsilon (\gamma) \left( \phi_{\mu \epsilon} \frac{d\phi_{\nu \epsilon}}{d\phi_{\delta \epsilon}} + \frac{d\phi_{\mu \epsilon}}{d\phi_{\delta \epsilon}} \phi_{\nu \epsilon} \right)
\end{align*}

\noindent Now, notice that ${d\phi_{\mu \epsilon}}/{d\phi_{\delta \epsilon}}$ is non-zero only if $\mu = \epsilon$. Denote by $\uparrow_{\mu \epsilon}$ the Kronecker delta function, which is 1 only if $\mu = \epsilon$, 0 otherwise. Then,
\begin{align}
\frac{d\chi_{\mu \nu}(\beta\gamma)}{d\phi_{\delta \epsilon}} &= d_\epsilon (\beta) d_\epsilon (\gamma) \left( \phi_{\mu \epsilon} \uparrow_{\nu \delta} + \uparrow_{\mu \delta} \phi_{\nu \epsilon} \right)
\end{align}

\noindent Next, 
\begin{align*}
\frac{d\xi_{\mu \nu}(\beta\gamma)}{d\phi_{\delta \epsilon}} &= \frac{d}{d\phi_{\delta \epsilon}} \sqrt{\left( \sum_{\alpha = 1}^{n} \phi_{\mu \alpha}^2 d^2_\alpha (\beta) \right) \left( \sum_{\tau = 1}^{n} \phi_{\nu \tau}^2 d^2_\tau (\gamma) \right)} \\
&= \frac{1}{2 \xi_{\mu \nu}(\beta\gamma)} \frac{d}{d\phi_{\delta \epsilon}} \left( \sum_{\alpha = 1}^{n} \phi_{\mu \alpha}^2 d^2_\alpha (\beta) \sum_{\tau = 1}^{n} \phi_{\nu \tau}^2 d^2_\tau (\gamma) \right) \\
&= \frac{1}{2 \xi_{\mu \nu}(\beta\gamma)} \left[ \sum_{\alpha = 1}^{n} \phi_{\mu \alpha}^2 d^2_\alpha (\beta) \frac{d}{d\phi_{\delta \epsilon}} \left( \sum_{\tau = 1}^{n} \phi_{\nu \tau}^2 d^2_\tau (\gamma) \right) \right.\\
&\quad \quad \quad \quad \quad \left. + \sum_{\tau = 1}^{n} \phi_{\nu \tau}^2 d^2_\tau (\gamma) \frac{d}{d\phi_{\delta \epsilon}} \left( \sum_{\alpha = 1}^{n} \phi_{\mu \alpha}^2 d^2_\alpha (\beta) \right) \right]
\end{align*}

\noindent Again, a term in one of the above summations is non-zero only if $\alpha$ or $\tau$ is the same as $\epsilon$. Thus,
\begin{align*}
\frac{d}{d\phi_{\delta \epsilon}} \left( \sum_{\alpha = 1}^{n} \phi_{\mu \alpha}^2 d^2_\alpha (\beta) \right) &= 2 \phi_{\mu \epsilon} d_\epsilon^2(\beta) \uparrow_{\mu \delta}
\end{align*}

\noindent Thus, 
\begin{align}
\frac{d\xi_{\mu \nu}(\beta\gamma)}{d\phi_{\delta \epsilon}} &= \frac{1}{\xi_{\mu \nu}(\beta \gamma)} \left[ \phi_{\mu \epsilon} d_\epsilon^2(\beta) \uparrow_{\mu \delta} \sum_{\tau = 1}^{n} \phi_{\nu \tau}^2 d^2_\tau (\gamma) + \phi_{\nu \epsilon} d_\epsilon^2(\gamma) \uparrow_{\nu \delta} \sum_{\alpha = 1}^{n} \phi_{\mu \alpha}^2 d^2_\alpha (\beta) \right]
\end{align}

\noindent This completes the calculation of derivatives. 

\section*{References}

\bibliography{references}

\end{document}